\documentclass[graybox]{svmult}

\usepackage{type1cm}        
\usepackage{makeidx}         
\usepackage{graphicx}        
\usepackage{multicol}        
\usepackage[bottom]{footmisc}

\usepackage{newtxtext}       %
\usepackage{newtxmath}       

\makeindex             

\begin{document}

\title*{Visual Analytics and Human Involvement in Machine Learning}
\author{Salomon Eisler and Joachim Meyer}
\institute{Salomon Eisler \at Tel Aviv University, \email{seisler@gmail.com}
\and  Joachim Meyer \at Tel Aviv University \email{jmeyer@tau.ac.il }}
%
%
\maketitle

\abstract*{Each chapter should be preceded by an abstract (no more than 200 words) that summarizes the content. The abstract will appear \textit{online} at \url{www.SpringerLink.com} and be available with unrestricted access. This allows unregistered users to read the abstract as a teaser for the complete chapter.
Please use the 'starred' version of the \texttt{abstract} command for typesetting the text of the online abstracts (cf. source file of this chapter template \texttt{abstract}) and include them with the source files of your manuscript. Use the plain \texttt{abstract} command if the abstract is also to appear in the printed version of the book.}

\abstract{The rapidly developing AI systems and applications still require human involvement in practically all parts of the analytics process. Human decisions are largely based on visualizations, providing data scientists details of data properties and the results of analytical procedures. 
Different visualizations are used in the different steps of the Machine Learning (ML) process.  The decision which visualization to use depends on factors, such as the data domain, the data model and the step in the ML process. In this chapter, we describe the seven steps in the ML process and review different visualization techniques that are relevant for the different steps for different types of data, models and purposes.}

\section{Introduction}
\label{sec:1}
The use of computers and sensors in practically all parts of life dramatically increased the amount of available data. Numerous visualization techniques and graphic tools have been developed to satisfy the human need to understand and analyze the information in the data. Keim \cite{Keim2002} and others suggested that for data mining (an old synonym of Data Science) to be effective, humans have to be included in the data exploration process to combine the flexibility, creativity, learning capability and general knowledge of the human with the enormous storage capacity and the computational power of today's computers. However, humans' cognitive abilities have not changed. A discrepancy exists between the large increase in the complexity of the data and human cognitive capacities. This can be seen as a problem of visual scalability,which is defined as the capability of visualization tools to effectively display large data sets in terms of either the number or the dimension of individual data elements \cite{Keim2006a,Liu2017}.

But what happens when humans do not have to 'see' the information, when the analysis and learning processes are done by a machine? The current huge demand for data scientists indicates that, even though ML and AI have become major tools for knowledge discovery with databases (KDD)\cite{Fayyad1996}, humans are still strongly involved in the process. 
To this end, humans need to gain knowledge about data properties and the results of analytical procedures. These "insights" rely to a large extent on the visual display of information, i.e., visualization. The choice of the visualization method and its implementation will depend on properties of the data, the problem for which the data is analyzed, the purpose of the analysis and other factors \cite{Meyer1997,Munzner2014}. In the following sections we will review the major tasks carried out by data scientists and the user interfaces and visual analytics related to them. 

\section{Overview: Visualizations used during the steps of the Machine Learning process}
\label{sec:2}
 
In this section we provide a high level overview of the steps of Machine Learning (ML) and discern when and why visualizations can be used to improve the process.  In the next section we provide details for each step.

\begin{figure}
			\centering
			\includegraphics[width=\textwidth]{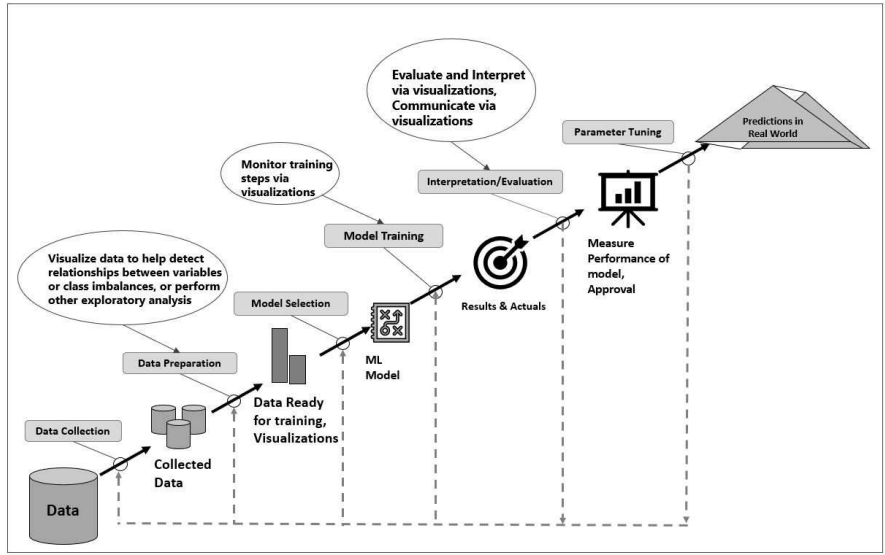}
			\caption{the 7 Steps of Machine Learning} \label{Sevensteps}
		\end{figure}

The seven steps of Machine Learning (ML) are Data Collection, Data Preparation, Model Selection, Model Training, Evaluation and Interpretation, Parameter Tuning and Prediction Making \cite{Mayo2018,Guo2017, Chollet2018}. They are similar to the process described by Fayyad for KDD \cite{Fayyad1996}. Fig.~\ref{Sevensteps} depicts these seven Steps of Machine Learning.

The analytical methods of ML usually require the data to have a certain form and structure. The process of converting the data into the required structure is called Data Preparation.  Transforming the data to a tabular format, removing or inferring missing values, removing outliers and anomalies and converting data to different types are examples of data preparation. Sometimes the techniques use categorical data, while others handle only numeric values. Usually, also, numerical values need to be normalized or scaled so that they are comparable \cite{FosterProvostandTomFawcett2013}. During this step, several tasks very frequently require to visualize data to help detect relevant relationships between variables or class imbalances, identify anomalies and outliers and perform other exploratory analyses \cite{Mayo2018}. Visualizations for this step should not be very different from the visualizations used in classical KDD, with the exception that they often involve large amounts of data, and the data scientist can encounter visual scalability issues.   

Model Selection is determined by the business or scientific question that has to be answered using ML, the type of data and its behavior. It is not infrequently that more than one model can answer the question. Therefore, one needs to understand the data to select the ML model, and visualizations may support this task. Again, as mentioned above, visualizations should not be very different from visualizations in traditional KDD. Even if visualizations are not used during Model Selection, visualizations may not be used, it is important to review this step, because the selected model may also determine the visualizations used in the next steps of the process.

The Model Training step is when the model learns using the data and computes its internal parameters.  This is done on a set of data that should be as ergodic as possible, so that when the model is applied to the testing data and to real world data, it will continue to provide good results. Here, visualizations are important, as they can help to monitor how the model converges to a solution.

The goal the Evaluation step is to assess the ML results rigorously to gain confidence that they are valid and reliable and that the model satisfies the original business goals before moving on \cite{FosterProvostandTomFawcett2013}. This is done, using a separate set of data, called test data, with the expectation of getting results, similar to those achieved during the training step. Visualizations can also support this step very efficiently.

The Interpretation step involves qualitative assessments. Various stakeholders have interests in the decision-making that will be accomplished or supported by the resultant models \cite{FosterProvostandTomFawcett2013}.  During this step, it is important to consider the comprehensibility of the model to stakeholders \cite{FosterProvostandTomFawcett2013}, and here visual analytics may be crucial. 

Once the model has been defined, the model building is done with a programming language, supported by ML frameworks. More details will be introduced in section 5.

The Parameter Tuning step refers to hyperparameter tuning, which is more an art than a technique. The objective is to improve performance through fine tuning of the number of training steps, learning rate, initialization values and distribution, etc. \cite{Guo2017,Mayo2018}.  The values of hyperparameters configure characteristics of the model and may highly impact the training performance. However, given the complexity of the model algorithms and the training processes, identifying a sweet spot in the hyperparameter space for a specific problem can be challenging (HyperTuner: Visual Analytics for Hyperparameter Tuning by Professionals - Li 2018). Dashboards of SPLOM, heatmaps and line plots  can be used to determine the optimal set of parameters.

The Prediction in the Real World step uses data that was not used during the training to do the real predictions and  actually use them \cite{Guo2017,Mayo2018}.  Prediction, or inference, is the step where the answers to the initial questions are received \cite{Guo2017}. In this step, visualizations are mainly used for monitoring the results over time and to detect if the ML model needs to be modified because of changes in the data behavior. This is less relevant for the data scientist and it is out of scope of this chapter.

In the next sections we will discuss the visualizations for Data Preparation, Model Selection, Training the model, Evaluation and Interpretation and Parameter Tuning steps in more detail.

\section{Visualizations in the Steps of the ML process}
\label{sec:3}

\subsection{Data Preparation}
\label{subsec:1}
As mentioned above, during Data Preparation one often wants to visualize data to help detect relevant relationships between variables or class imbalances, extract anomalies and outliers and to perform other exploratory analyses \cite{Mayo2018}.
With the huge increase in the number of available observations and the number of features (variables) for each observation, it has become harder to provide clear and easy to understand visualizations. One way to overcome the visualization scalability problem is to use automatic analysis methods to extract potentially relevant visual structures from a set of candidate visualizations and rank the visualizations in accordance with a specified user task \cite{Tatu2010}. Once a manageable number of potentially useful visualizations are identified, the data scientist can start interactive data analyses \cite{Tatu2010}. 

From a data perspective, the most popular automatized available tool for simplifying visualizations and for providing a better understanding is Dimensionality Reduction (DR). DR has become a core building block in the visualization of multidimensional data \cite{Sacha2017}. Examples of DR algorithms can be found in \cite{Etemadpour2015,Holzinger}. Studies,  \cite{Etemadpour2015,cuadros2007point}, have used scatter plots to evaluate users' perceptions for different DR algorithms on four different data-sets and four techniques to achieve DR.  Not surprisingly, it was shown that performance depends on data characteristics \cite{Etemadpour2015}, and the density and surrounding information affects the perception of clusters \cite{sedlmair2012taxonomy}. Performance will also be affected by characteristics of the individual user's cognition and perception \cite{bak2005effects,Green2008}. 

\begin{figure}
			\centering
			\includegraphics[width=\textwidth]{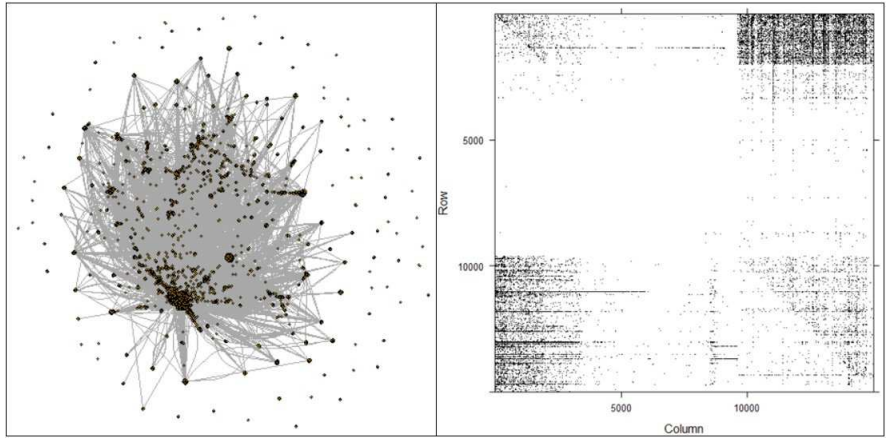}
			\caption{Left, graph visualization; right, adjacency matrix using the R igraphs package  } \label{Graph_red}
		\end{figure}

It is also possible to apply DR to Network data that is visualized through graphs. A first step is to convert the graph data into a sparse matrix (called an adjacency matrix) that can be easily visualized. The second step is usually a matrix reorganization to reduce the graph to smaller graphs by partitioning its nodes into mutually exclusive groups \cite{burkhardt2017graphing}. The Laplacian matrix transformation is the most common example of this type of graph partitioning. The premise in the reordering of the adjacency matrix is to align the non-zero values close to the diagonal of the matrix, reducing the geometric distances between vertices, which results in a simpler visualization of graph partitioning \cite{burkhardt2017graphing}. Fig.~\ref{Graph_red} shows graph visualization of 15000 nodes from an Email network, based on traffic data collected for 112 days \cite{Ebel}. 

\subsection{Choosing a model}
\label{subsec:3}

ML models are either unsupervised or supervised learning models, depending on the type of problems they are intended to solve. With supervised learning, the machine ("the learner") receives target information, along with a samples collection  \cite{FosterProvostandTomFawcett2013}.  The machine must develop a solution, which will match the target with the sample data.  In unsupervised learning, target information is not included. The machine is left to reach its own conclusions about the common properties that exist in the samples collection \cite{FosterProvostandTomFawcett2013}.

Numerous supervised and unsupervised ML methods exist and require different visualization techniques. It is beyond the scope of this chapter to present an exhaustive review of all types of visual analytics for all methods, and we will focus only on some popular ones.

The visualization techniques to be used for a specific Machine Learning model, as mentioned above, depend on the combination of the data-types and the selected model \cite{Munzner2014}. In general, the types of visualizations used in ML are not complex or particularly innovative. To the contrary, they are usually simple and intuitive: line plots, scatter plots, heat-maps, contour plots,  hierarchical trees and several combinations of the above. The main differences between the visualization techniques are in the data elements that are visualized and the methods used to transform them. 
There are some exceptions, such as image processing and data graphs. In face recognition, the visualizations can be the arrays of face images. With network data, the visualizations are graph representations of the network, using different layouts of the data points (either in 2D or 3D). Sometimes symbols are used to consolidate several data points, with specific meaning related to the type of connections of the group of data points points to reduce the visual clutter. 
Below is a short overview of the different ML models and their related visual analytics. 

\subsubsection{Supervised models}  
\runinhead{Classification trees:} Users either want to edit the tree (grow, prune or optimize it), use the tree (classification) or analyze it (data exploration). Typically, users often switch between the edit and analysis process. For each task, one can identify important elements and extract requirements \cite{van2011baobabview}. Several examples of classification trees models were cited in \cite{Holzinger}, and \cite{van2011baobabview} presents a good example of an interactive interface with a decision tree that supports editing, classifying and exploring the tree.

\begin{figure}[t]
\sidecaption[t]
\includegraphics[scale= 0.3]{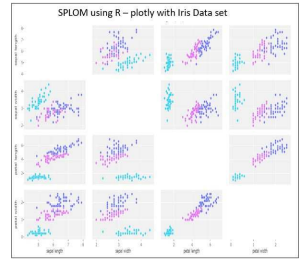}
\caption{SPLOM plot using R and plotly package - \cite{Plotly}  \texttt{ } }
\label{SPLOMS}     
\end{figure}

\runinhead{Regression models:} These models can be used to isolate the relationship between outcome and explanatory variables, while holding other variables constant. Visualizations are important when interacting with this kind of models, because it is possible to visually represent these relationships in an easy to understand way with simple scatter plots.  When the relationship between an explanatory variable and the response depends on multiple regression coefficients, the model's fit is more readily understood with a visual representation than by looking at a table of regression coefficients \cite{breheny2013visualization}. Examples of classification regression models are shown in \cite{Holzinger}.
Fig.~\ref{SPLOMS} shows a scatter plot matrix (SPLOM) that can be used to quickly explore distributions (clustering - unsupervised) and relationships (regressions -  supervised), based on the known Iris Data set and created with R.

\begin{figure}
			\centering
			\includegraphics[width=\textwidth]{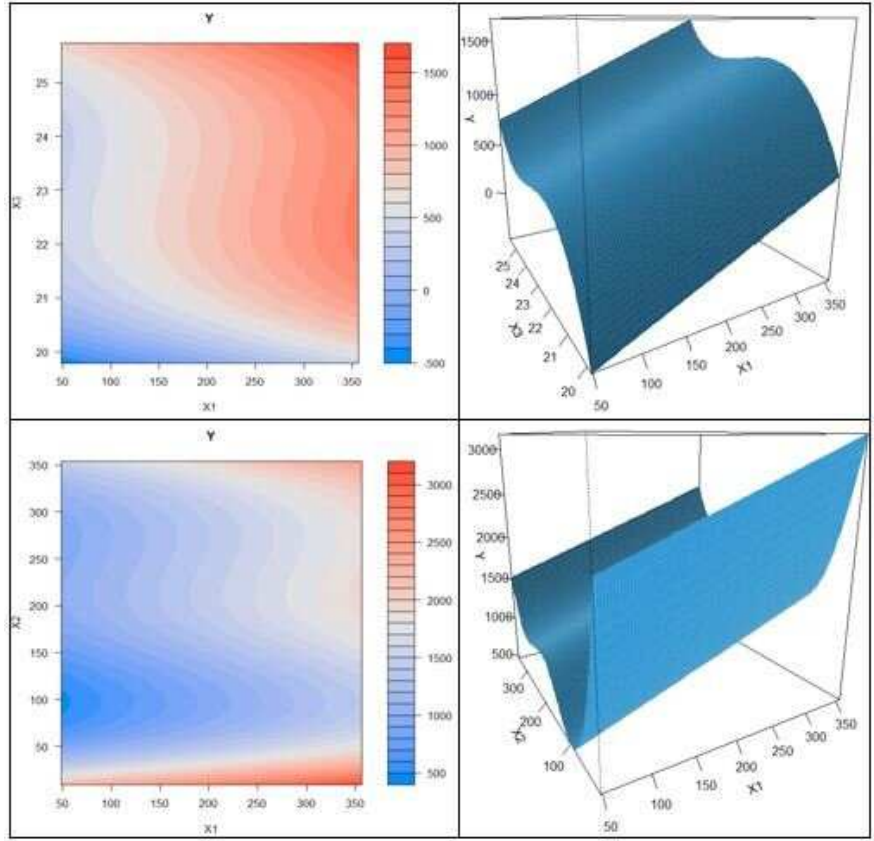}
			\caption{ Representations of the regression surface as a function of X1, X2 and X3, using synthesized data where Y = 22 + 3X1 + 2X2 + 3X3. Left: Filled contour plots. Right: Perspective plots. Using R and code from \cite{breheny2013visualization}.} \label{Contours}
		\end{figure}

Cross-sectional plots, contour plots and 3D representations of the regression surface \cite{breheny2013visualization} are helpful visualizations for Regression Models. Fig.~\ref{Contours} depicts examples of this type of visualizations.

\runinhead{Bayesian networks (BNs):} These plots, also known as belief networks \cite{ben2008bayesian}, are related to the family of probabilistic graphical models (GMs). Their graphical structures are used to represent knowledge about an uncertain domain \cite{ben2008bayesian}. Visualizations of this type of models are similar to decision tree visualizations. An example of a tool that visualizes BNs is BayesViz's \cite{chiang2005visualizing}. This tool presents the inferred network with edges, colored per correlation coefficient, and colormap tables, representing the conditional relationships between the values of parent and child nodes. 

\runinhead{Decision Tree Inducers:} Decision trees are constructed with inducers (also called classifiers). A decision tree inducer is basically an algorithm that automatically constructs a decision tree from a given (training) data-set. Visualizations here are practically the same as in the decision trees above. Several models of this type exist: ID3, C4.5, CART, CHAID, QUEST, CRUISE and many others \cite{Holzinger} \cite{Ratnaparkhi2017} .

\runinhead{Support Vector Machines (SVM):} These models are used for both classification or regression challenges. SVMs are the only linear models which can classify data that is not linearly separable \cite{ben2008bayesian}.  Visualizations are used in SVM to understand the decision boundary in the space of input variables. This decision boundary is estimated from available training data, but is intended for classifying future input samples \cite{chiang2005visualizing}. Examples of SVM models are shown in \cite{Holzinger}. In  \cite{ben2008bayesian,chiang2005visualizing} it is possible to see a comparison of visualizations of the decision boundary for linear and non-linear SVM model.

\subsubsection{Un-Supervised models}  
\begin{figure}[t]
\sidecaption[t]
\includegraphics[scale=.55]{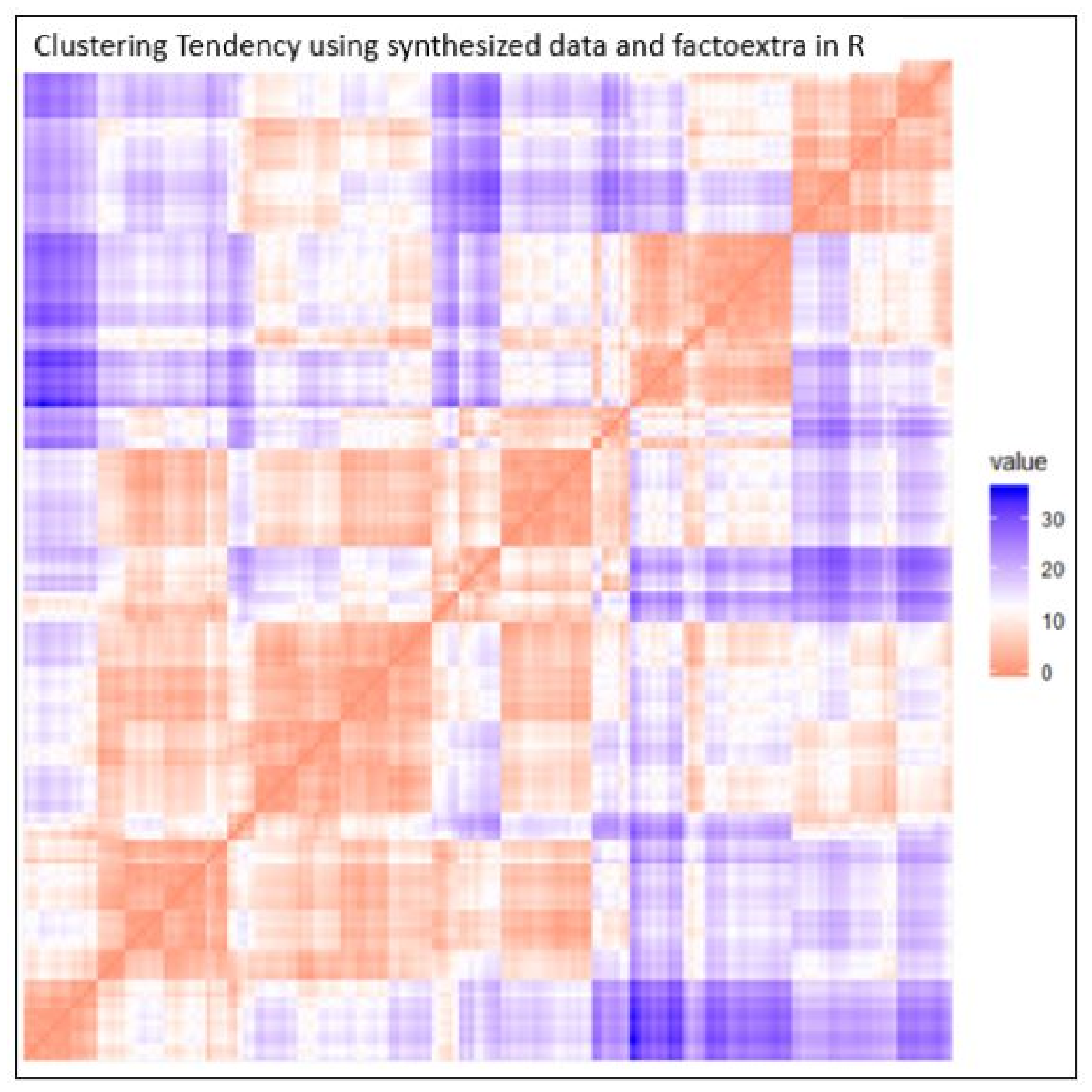}
\caption{ Clustering tendency  \cite{kassambara2017practical} is detected in a visual form by counting the number of square shaped dark blocks along the diagonal in the image. The data set used is synthesized data shown in Fig.~\ref{density}. Red: high similarity (ie: low dissimilarity) , Blue: low similarity.  \texttt{ } }
\label{clust_tend}     
\end{figure}

\begin{figure}
			\centering
			\includegraphics[width=\textwidth]{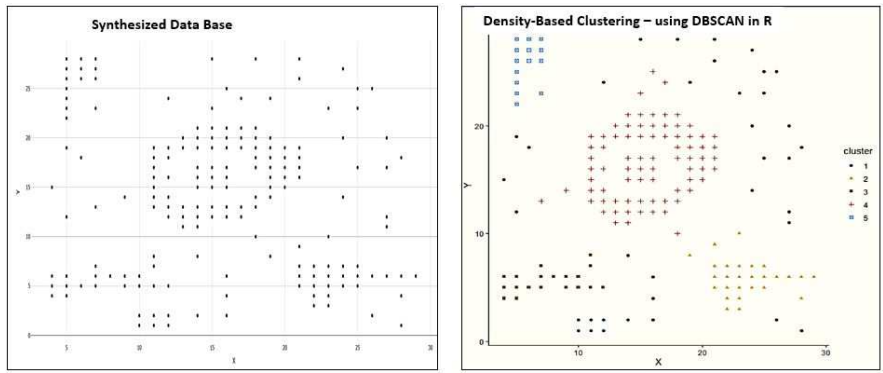}
			\caption{Density based clustering can help to find clusters with different shapes and sizes from data containing noise and outliers \cite{kassambara2017practical}} \label{density}
		\end{figure}

\runinhead{Clustering:} Before applying any clustering algorithm to a data set, one first has to assess the clustering tendency (to understand whether the data set has a natural clusters or not) (Fig.~\ref{clust_tend}  \cite{kassambara2017practical} ).  The classification of objects into clusters requires methods for measuring the distance or the (dis)similarity between the objects. Visualizations can help to expose and analyse both, the clustering tendency and similarity distances in the data \cite{kassambara2017practical}. 
Density based clusters \cite{ester1996density} are an additional way to visualize possible clusters. The main reason why it is possible to recognize clusters in Fig.~\ref{density} is that the density of points within each cluster is clearly higher than the density outside the cluster \cite{ester1996density}.

The percentage of data points above the Similarity-Dissimilarity line (PAS) shows the expected accuracy of the classifier, using a particular feature set \cite{arif2012similarity}. The Similarity-Dissimilarity visualization for a high dimensional feature space can provide very important information that can later help in the development of the models \cite{arif2012similarity}.

Hierarchical Clustering is another type of a clustering analysis method that is well supported by visual analytics. Hierarchical clustering is an algorithm, which is intended to create a hierarchy of clusters. The last layer of the hierarchy is a group of clusters, where each cluster is separate from the others, and the objects within each cluster are very similar to each other.
Strategies for hierarchical clustering generally fall into two types: agglomerative and divisive. Agglomerative is a "bottom-up" method: each observation starts in its own cluster (leaf), and pairs of clusters are merged as one moves up the hierarchy until there is just one single big cluster (root) \cite{kassambara2017practical}. Divisive is a "top-down" approach: all observations start in one cluster (the root), and splits are performed recursively down the hierarchy.

\begin{figure}
			\centering
			\includegraphics[width=\textwidth]{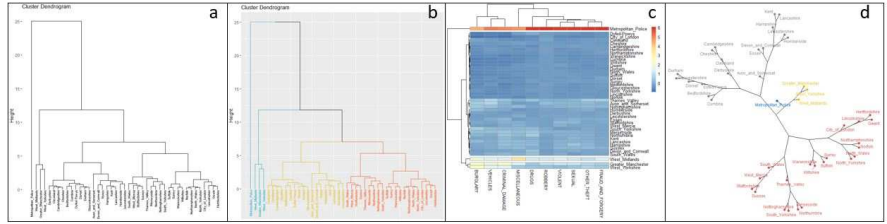}
			\caption{a) Simple dendrogram, b) dendrogram with 4 groups, c) Heat Map using package pheatmap
d) Phylogenic dendrogram.  Visualizations were prepared with R.
 } \label{dendogram}
		\end{figure}

The visual representation of hierarchical clustering is a tree-based visualization of the objects, which is also known as dendrogram \cite{kassambara2017practical}. It can also be used to analyze network data.  Fig.~\ref{dendogram} depicts four visualizations of the same data set (Offences recorded by the police in England and Wales by offence and police force area for 2001/02, from https://www.gov.uk/government/statistics/historical-crime-data), with dendrograms and a heat map using R.

\runinhead{Graphs - Network Data:} A network consists of a collection of entities and the connections or relations between them. It can be visually represented as a graph, with vertices representing entities and edges representing their relationships \cite{o2005analysis}. This is an intuitive representation, because of the close similarity between the real world and the visualization. When the data scientist explores the data via visualization, he/she is exploring almost the actual network. 
Networks are a special case, where the data scientist can use different ML techniques, while exploring the graph representing the network.  These include routines for clustering, decomposition, random graph generation, statistical analysis, and calculation of network distances \cite{cherkassky2010simple}. 
Graphs need a separate approach for almost all the 7 steps of the ML process. An example that specializes in graph data with specific graph visualizations is GRAPHVIS, which supports interactive techniques, such as brushing, linking, highlighting, as well as semantic zooming.

\subsubsection{Advanced Methods}
\runinhead{Deep Learning} includes a wide range of techniques, including Neural Learning Networks, Genetic Algorithms \cite{arif2012similarity}, Convolutional Neural Networks, Recurrent Neural Networks, Deep Belief Networks, etc. One characteristic of these methods is that it is hard to understand and interpret the underlying rules or mechanisms that produce the predictions, i. e., the methods are black boxes. These methods can be supervised or un-supervised. As we will see later, for the adoption and use of the model it is often crucial to assure that one can interpret and explain the model, and this can be done using visualizations. One way of trying to interpret a neural network model is to create heat maps of the cells of the actual neural network.  \cite{o2005analysis} present an example of a visualization for a neural machine that translates it. There, the x-axis and y-axis of a heat map plot correspond to the words in the source sentence (English) and the generated translation (French), respectively. Each pixel shows the weight ij of the annotation of the j-th source word for the i-th target word, which correspond to the cells of the neural network.

\runinhead{Ensemble Models} are models that make predictions, based on a group of different models. While deep learning models are more appropriate in fields, such as image recognition, speech recognition, and natural language processing, tree-based ensemble models frequently outperform standard deep models with structured data where features are individually meaningful \cite{chen2016xgboost}. Visualizations of these decision trees are relatively easy to understand, as they show a hierarchical view of the step decisions, made by the classification model. 

\subsection{Training the model}
\label{subsec:4}

The goal of training is to enable the machine to learn the data, so that the answers to questions or the prediction are as correct as possible. If the processing is too heavy and takes too long, independently from the ML framework used,  trained models need to be saved in a file and afterwards restored to compare the model with other models, to test the model on new data  or for checkpoints. The saving of data is called serialization, and restoring the data is called deserialization. Supporting this task with visualizations can be very helpful to the data scientist. An example of a framework that enables this feature is TensorFlow with TensorBoard  (its suite of visualization tools) \cite{abadi2016tensorflow,Brain,Ganegedara2018}.

\begin{figure}[t]
\sidecaption[t]
\includegraphics[scale=0.4]{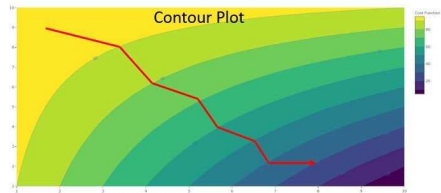}
\caption{Contour plot illustration using R. The Red line shows the gradient descent convergence.  \texttt{ } }
\label{contourplot}       
\end{figure}

One also often needs to monitor the algorithm iterations to verify convergence and to evaluate the results. There are several types of visualizations that enable the data scientist to manage this process. For example, the progress of gradient descent on a test surface can be visualized for the all the steps  \cite{Heindl2018}.  Sometimes it is useful to display the three-dimensional data in two dimensions, using contours or color-coded regions \cite{vanderplas2016python}. Fig.~\ref{contourplot} depicts a contour plot for gradient descent. The red arrow shows the convergence of the process to the minimum of the cost function.

\subsection{Evaluation and Interpretation}
\label{subsec:5}

\subsubsection{Evaluation}

\begin{figure}
		\centering
		\includegraphics[width=\textwidth]{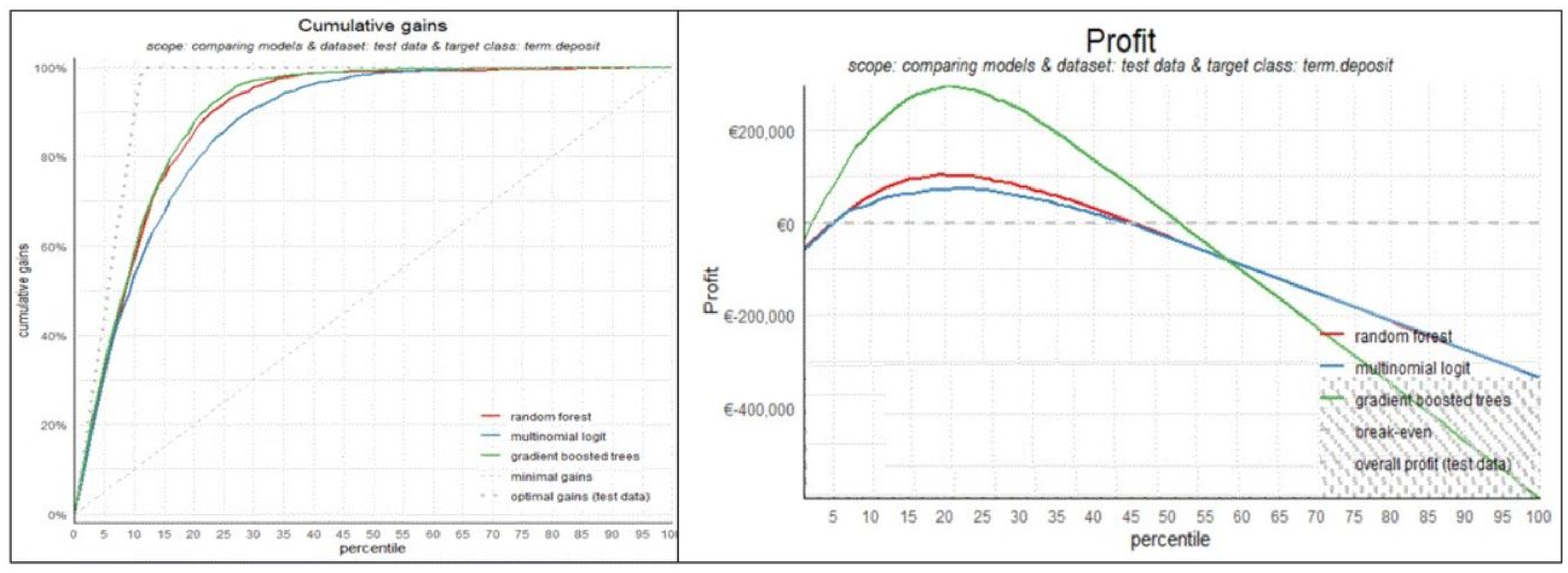}
		\caption{ Cumulative Gains and Profit plot for classifiers   }
		\label{Profits} 
		\end{figure}

Human interaction is very important in the evaluation step of the process. The right data, combined with the right data science techniques, will help to identify the models that optimize a cost criterion. However, only humans can decide on what is the best criterion  \cite{FosterProvostandTomFawcett2013}. 
One strategy for making decisions is to rank a set of cases by scores, and then take actions on the cases at the top of the ranked list. This can be achieved using profit curves, which consider costs/benefits related to true positives, false positives, true negatives, and false negatives. Profit curves are appropriate when the conditions under which a classifier will be used are known with high certainty. A profit curve can help optimize overall profit and help select the best model and predicted probability threshold \cite{Lai2016}.  Fig.~\ref{Profits} shows an example of cumulative gains and profit curves for three classifiers, using R package modelplotr based on the Bank Marketing Data Set \cite{Moro2014}.

\begin{figure}[t]
\sidecaption[t]
\includegraphics[scale=0.2]{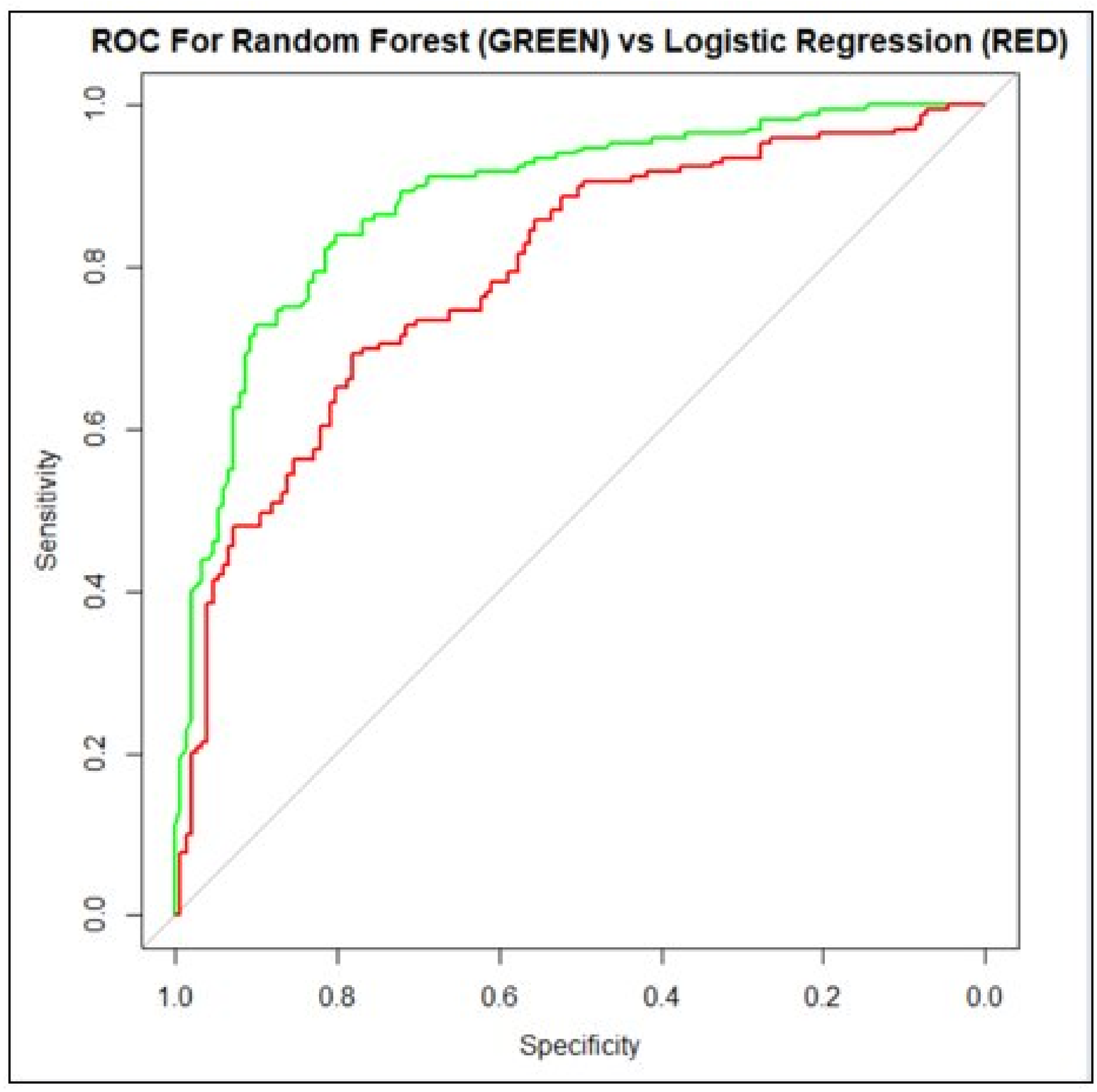}
\caption{ ROC curve. The diagonal line x=y represents random performance, comparing 2 classification models.   \texttt{ } } 
\label{ROC}       
\end{figure}

Different evaluation methods should be used when the conditions under which the classifiers are used are uncertain or unstable.  One such method is the Receiver Operating Characteristics (ROC) curve. ROC curves can serve as the basis for performance measurements in classification problems at various thresholds settings. ROC is a probability curve and Area Under the Curve (AUC) represents the degree of how much the model is capable of distinguishing between classes. The higher the AUC, the better the model predicts 0s as 0s and 1s as 1s or distinguishes between positives and not positives \cite{Narkhede}.   Fig.~\ref{ROC} depicts an example of a ROC curve, the diagonal line x=y represents random performance, using the wine quality data set from https://archive.ics.uci.edu/ml/data-sets/wine+quality  and the pROC R package, code from https://www.kaggle.com/milesh1/receiver-operating-characteristic-roc-curve-in-r.
Another, more intuitive visualization for model evaluation is the 'cumulative response curve'.  Cumulative response curves plot the true positive rate, which is the percentage of positives correctly classified, as a function of the percentage of the population that is targeted (x axis) \cite{FosterProvostandTomFawcett2013}.

\begin{figure}[t]
\sidecaption[t]
\includegraphics[scale=0.2]{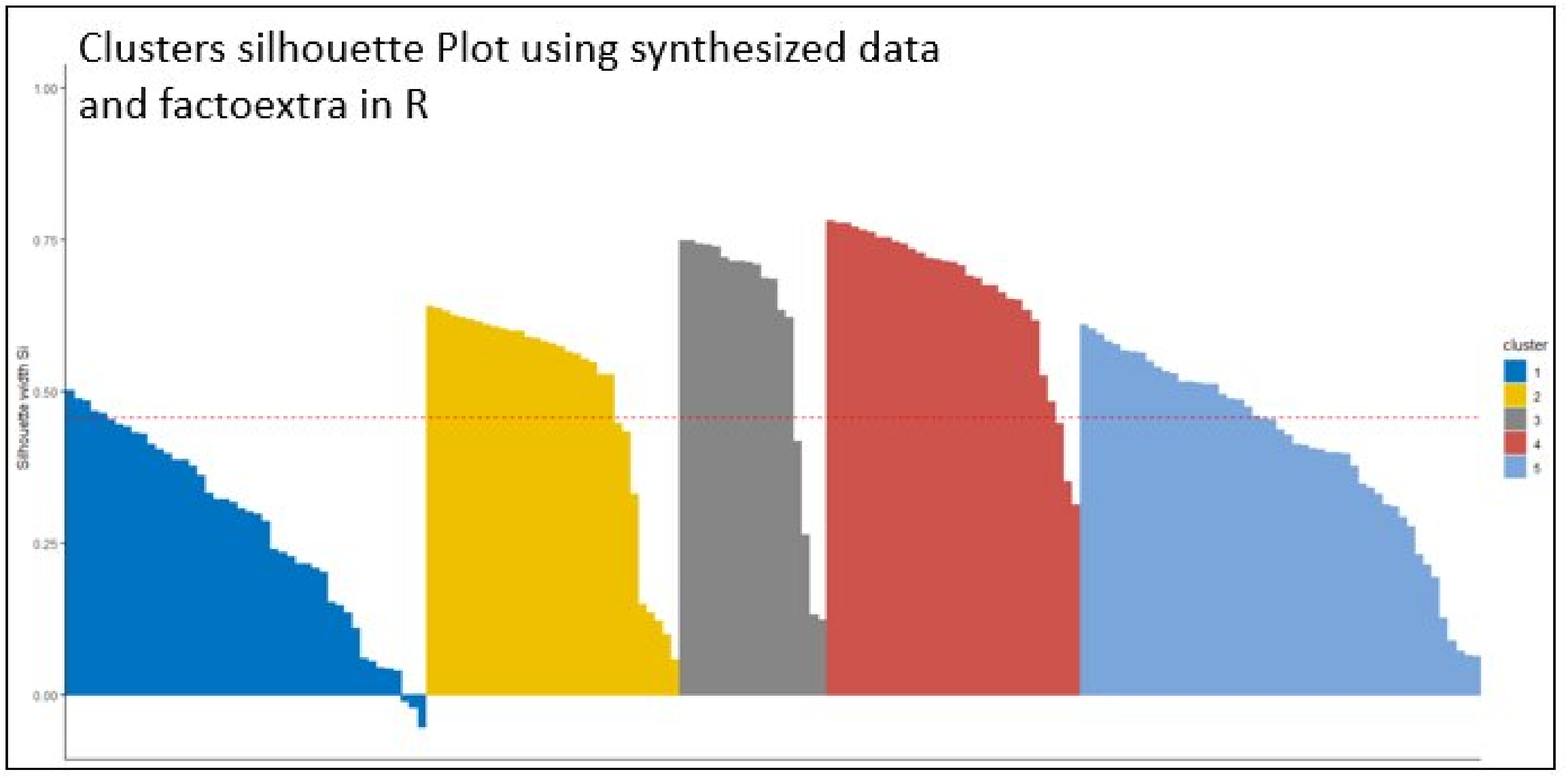}
\caption{ \cite{kassambara2017practical} Clusters silhouette plot, Silhouette width is also an estimate of the average distance between clusters. Values are between 1 and -1, with a value of 1 indicating a very good cluster.
   \texttt{ } } 
\label{silhoutte}       
\end{figure}

Another metric that can be useful for evaluating a clustering model is the Silhouette Coefficient (Si) \cite{Rousseeuw1987}.  It is a visualization method for interpreting and validating the consistency within clusters of data. A value of Si close to -1 indicates that the object is poorly clustered. The silhouette plot displays a measure of how close each point in one cluster is to points in the neighboring clusters. It thereby provides a way to visually assess parameters, such as the number of clusters.  Fig.~\ref{silhoutte} shows the silhouette plot of a k-means clustering, using the synthesized data set shown in Fig.~\ref{density} .

Typically, data scientists will perform several training processes to be able to compare model performance. The visual comparison of model performance for several experiments can result in heavy cognitive load for the data scientist.  A couple of examples that support this visual process are Squares \cite{ren2016squares} and iVisClassifier \cite{choo2010ivisclassifier}. Squares has been used for displaying the performance of a classifier trained on a handwritten digits data-set.   iVisClassifier includes dimensionality reduction techniques, scatterplots and parallel coordinates to support examination of model behavior.
 
 In ML, a confusion matrix is commonly used to present the accuracy of a learning algorithm. It is possible to compare model performance, showing the evolution of the confusion matrix over the training steps   \cite{liu2017visual}. Its visual representations are stacked plots that reveal the changes of True Positives and False Positives for each feature over iterations \cite{liu2017visual}.

\subsubsection{Interpretation}

One of the most debated topics in deep learning is how to interpret and understand a trained model, particularly in the context of high-risk industries, such as healthcare \cite{SHAIKH}. The Interpretation step includes interpreting the discovered patterns and possibly returning to any of the previous steps, as well as visualization of the extracted patterns, removing redundant or irrelevant patterns and translating the useful ones into terms users can understand \cite{Holzinger}.  The visualizations in this step are critical, as they can be crucial in getting the organization's trust to accept and adopt the specific ML model and algorithms \cite{Ribeiro2016}.  Visualization methods provide the necessary means to simultaneously analyze the huge amount of information hidden in a deep learning  neural network \cite{Bischof}. 
When interpreting deep learning networks, such as Neural Networks, Convolutional Neural Networks or Deep Belief Networks, it  is possible to divide the visualization methods into three classes \cite{SHAIKH}:

\runinhead{Preliminary methods:}  Simple methods, which show the overall structure of a trained model.  These methods just show a diagram of the neuron's connections. In \cite{Theano2017} an example of such a method is shown for the overall structure of a model using the Theano framework.

\runinhead{Activation based methods:} Beyond the model definitions and the quantitative analyses, there is a need for qualitative comparisons of the solutions learned by deep learning models. It is possible to find good qualitative interpretations of high-level features represented by such models. Such interpretations are possible at the unit level, by visually reviewing and comparing visualizations of units from  different hidden layers. This is simple to accomplish and results are consistent across various techniques \cite{erhan2009visualizing}.  

\runinhead{Gradient based methods:} These methods tend to manipulate the gradients that are formed from a forward and backward pass while training a model. Saliency maps are a visualization technique based on gradients, to understand and visualize the nonlinearities embedded in feed-forward neural networks \cite{morch1995visualization}. Examples can be seen in \cite{simonyan2013deep}. Color segmentation is used, because the saliency map might only capture the most discriminative part of an object, and saliency thresholding might not be able to highlight the whole object. Therefore, the map with the thresholding was propagated to other parts of the object, which was achieved  using the color continuity cues \cite{simonyan2013deep}.

\begin{figure}
			\centering
			\includegraphics[width=\textwidth]{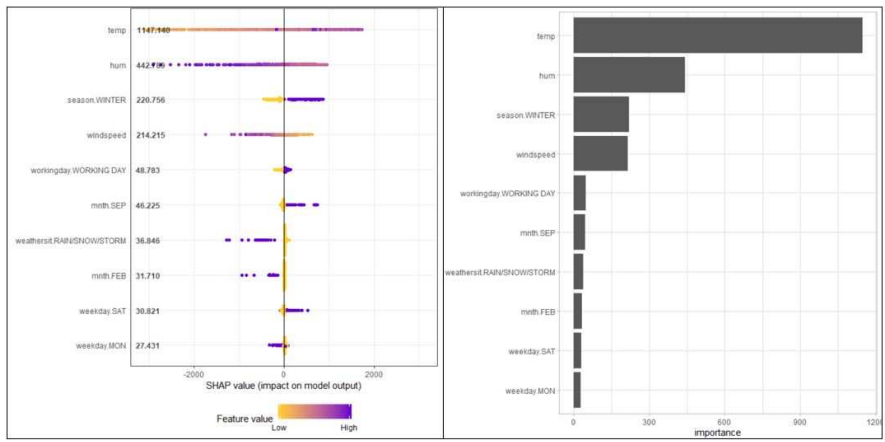}
			\caption{\cite{lundberg2018consistent} SHAP plots for a rental bike data-set \cite{Fanaee} using R code from \cite{Casas2018}. Left - SHAP summary plot of a 10-feature model. Right - importance plot, based on SHAP, using a Classical bar chart, showing the importance of the predictors.} \label{shap_val}
		\end{figure}

\runinhead{SHAP (SHapley Additive exPlanations)} is currently the only explanation method based on a solid theory with reasonable foundations \cite{molnar2018interpretable}.  It is a unified approach to explain the output of any machine learning model. SHAP connects game theory with local explanations, representing the only consistent and locally accurate additive feature attribution method based on expectations \cite{Lundberg2017}\cite{lundberg2018consistent}. All features can be contributors and try to predict the task, which is the game. 
The reward is the actual prediction minus the result from the explanation model \cite{Ma2018}. In SHAP, feature importance is assigned to every feature, which is equivalent to the mentioned contribution  \cite{Ma2018}. The impact of features is frequently plotted with bar charts to represent global feature importance, or with a partial dependence plot to represent the effect of changing a single feature \cite{friedman2001elements}.  SHAP summary plots replace typical bar charts of global feature importance, and SHAP dependence plots provide an alternative to partial dependence plots that capture interaction effects better \cite{lundberg2018consistent}. Fig.~\ref{shap_val} depicts a SHAP summary plot of a 10-feature model.  The y-axis indicates the variable names, in order of importance from top to bottom. The value next to them is the mean SHAP value. On the x-axis is the SHAP value, which indicates the change in log-odds. From this number, it is possible to extract the probability of success. Gradient color indicates the original value for that variable. Each point represents a row from the original data-set \cite{Casas, Casas2018}

\begin{figure}
			\centering
			\includegraphics[width=\textwidth]{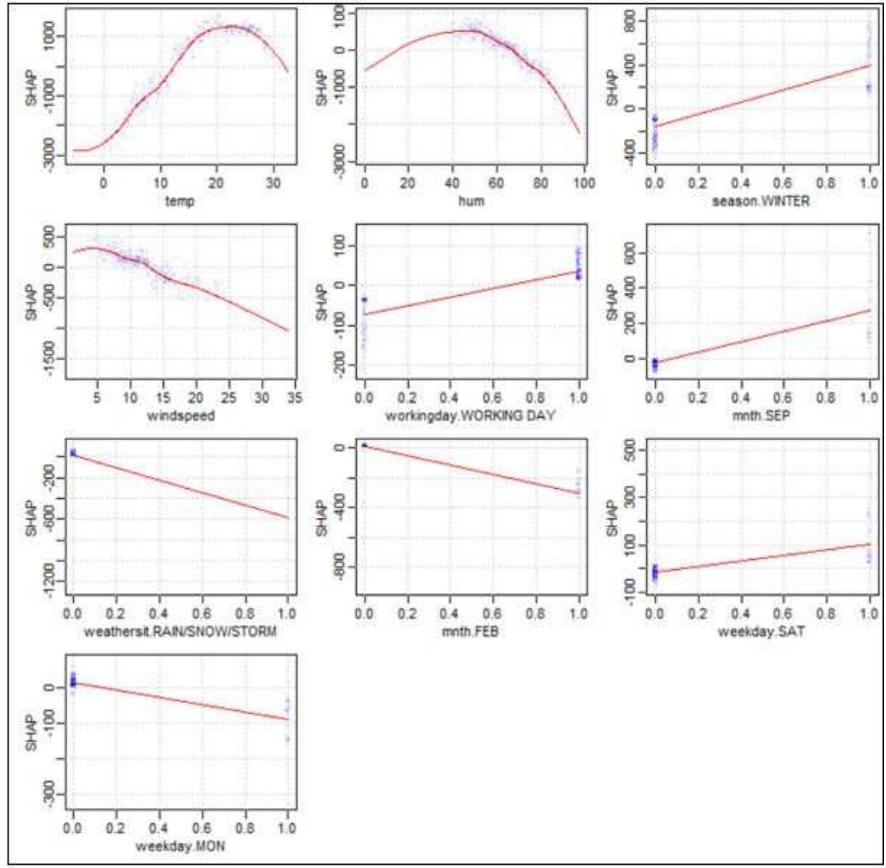}
			\caption{SHAP dependence plots for each of the 10 features from the rental bike data-set  \cite{Fanaee}, using R code from \cite{Casas2018}. } \label{shap_dependency}
		\end{figure}

Fig.~\ref{shap_dependency} depicts SHAP interaction dependence plots, which use the SHAP value of a feature for the y-axis and the value of the feature for the x-axis to present how the feature's attributed importance changes as its value varies. SHAP dependence plots capture vertical dispersion, due to interaction effects in the model. These effects can be visualized by coloring each dot with the value of an interacting feature. 

As mentioned above, graphs need a separate treatment for almost all the 7 steps of the ML process. There are two main entities that are usually considered when exploring graph data: Triplets and Motifs. Computing triangles in graphs has wide applications in network analysis, for identifying dense subgraphs, and for uncovering hidden thematic layers \cite{burkhardt2017graphing}.  Triangle computations help to visualize clusters in graph and support the data exploration during data preparation.
Motifs are frequently recurring patterns of basic structural elements that occur in graphs \cite{dunne2013motif}. They are small, local, patterns of interconnections that occur throughout a network with significantly higher probability than in random networks.  Motif detection is helpful for interpreting the graph.  One way to detect and to simplify motifs visualizations is by replacing motifs in the network with easily understandable glyphs \cite{dunne2013motif}. 

 A similar approach, which goes beyond motifs is identifying and visualizing clusters of motifs, called modules in \cite{li2015modulgraph}, in the graph and then visualize the relationships between the modules in a hierarchical way \cite{li2015modulgraph}. This reveals graph patterns for each module and allows users to gain a better understanding of the structure of the graph. ModulGraph \cite{li2015modulgraph} graph visualization can show the relations between the communities of motifs, and the different kinds of communities.

Another element when exploring a graph is to use graphlet frequencies to analyze the topological similarity of its sub-graphs using graph kernels \cite{kwon2017would}.  In simple terms, the graphlet frequency vector of a graph is like the feature vector of the graph  \cite{kwon2017would}.  If the graphlet frequency vector are not similar then the layout of each graph will look different. 

\subsubsection{Hyper-parameters tuning}
Today's hyperparameter tuning processes are highly empirical, using rules-of-thumb. They are "human driven", as they are performed manually by the data scientists. The ML tuning step has been described as a "a project involving multiple experiments, which may last several hours or days".  The number of hyperparameters differs between models, but there may often be more than 12 parameters.  
The optimizing algorithm, dropout rate, the number of layers, the width of each layer are hyperparameters that are commonly tuned.  For supervised models, the key performance metrics (KPIs) are usually accuracy, precision, recall, and ROC curves. Additional KPIs are the learning curve (how steep is the initial step and when it gets constant), training loss vs. validation loss (to detect overfitting) \cite{li2018hypertuner}.
To create visual analytics for hyperparameter tuning, it is necessary to capture and efficiently store the received KPIs, together with the employed parameters, per each tuning round. In this regard, this is similar to the traditional BI approach, where the data is stored and then visualized, using dashboards with combinations of SPLOM, heatmaps and line plots for different combinations of the KPIs and hyperparameters. Such an approach was described by Tian \cite{li2018hypertuner}.

\subsubsection {Summary}
\begin{figure}
\centering
\includegraphics[width = 9cm]{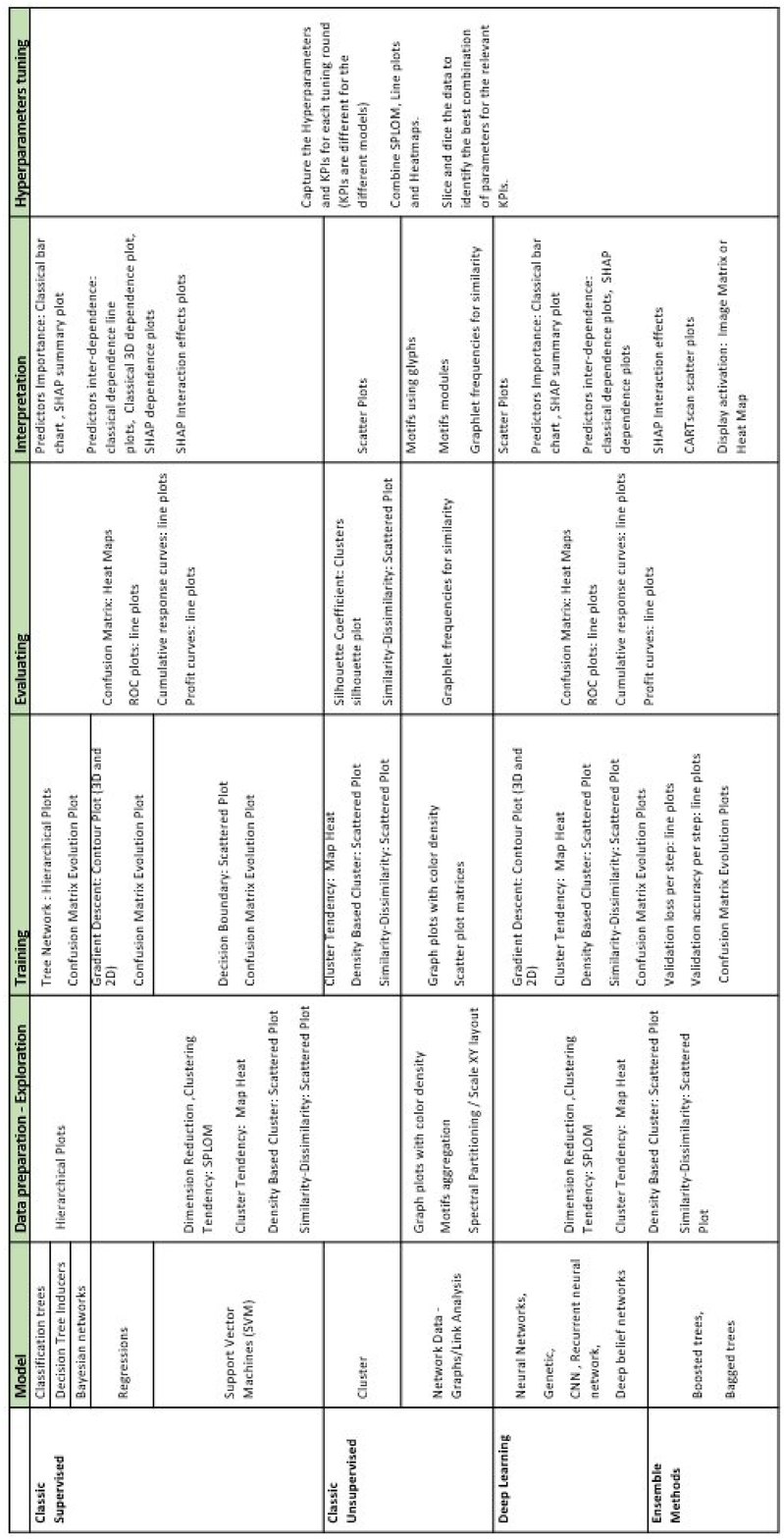}
\caption{Taxonomy - visual analytics and the 7 steps of Machine Learning   \texttt{ } }
\label{Taxo}     
\end{figure}

In Fig.~\ref{Taxo} we summarize the visualization techniques methods using  a mapping of the techniques related to 6 of the 7 Machine Learning steps: Data Preparation, Training, Evaluation, Interpretation and Hyper-parameters tuning for the following types of models: Classic Supervised and Unsupervised models, Deep Learning models and Ensemble models.

\section{User Interfaces and Frameworks}

As mentioned in section 2, once the model has been defined, it is necessary to build it.  Model building is  done using a programming language, such as Python, R, C++, JavaScript, or Java (see \cite{Heath2019,Nicholson,Shaleynikov2018} for a more complete list), supported by well-known ML frameworks such as TensorFlow, Torch, PyTorch, Jupyter Python, etc. (see \cite{Heath2019,Nicholson,Shaleynikov2018} for a more complete list). ML frameworks are interfaces, libraries or tools which allow developers build machine learning models easily and quickly, without needing to code all the underlying algorithms. ML frameworks have collections of pre-built, optimized components. They come with user interfaces that attempt to make it easier to understand, debug and optimize the ML programs.

Not all the ML frameworks have the same level of interactive features. Some of them only provide a coding platform for the data science practitioner. Others can turn into an end-user application for the expert analyst, including interactive visualization features, such as zooming and filtering different regions of the display, switching between types of visualizations and moving graphs via drag and drop. This is the mainly the case for interactive visualizations with graphs, such as GRAPHVIZ (www.graphviz.org) and D3VIZ from Theano  \cite{Theano2017}. 
As the development of interactive visualization frameworks for ML is trying to catch up with the explosion of ML development, many applications and tools are appearing in this domain.  On the other hand, these visualization tools mainly focus on specific steps of the ML process or on specific models. Examples are BOOSTVis, which is a visual diagnosis tool to help experts analyze and diagnose the training process of tree boosting \cite{liu2017visual}, or iVisClassifier\cite{choo2010ivisclassifier} for face recognition.

\section{Discussion}
One of the big challenges of ML adoption in organizations is understanding and interpretation \cite{Hodler2019}.  Trust is hard to get when the ML algorithm is a 'black box' \cite{Ribeiro2016}. The areas of explainability and interpretability are still emerging. Whenever a new ML model is proposed, questions arise regarding the data used and how the model works and how it will impact the current processes.  Visualizations are one of the best ways to interpret and explain the data and the models.

Questions, such as what was the data used to train the model, and why was this data and model combination used, often do not have the straightforward answers one might expect. To partly address this problem, data lineage methods, using visualizations, can be employed to explain how the data was changed and transformed \cite{Hodler2019}. 

In many academic fields, algorithms showing high explanatory power are often assumed to be highly predictive \cite{Wong2019}, but this is not always the case. There are many situations where building the best predictive model differs from building the best explanatory model \cite{Wong2019}, and modeling decisions often result in trade-offs between the two objectives.  When the only objective is to get the best prediction, the goal is quite clear: to find a systemic function that can be treated as a black box, which will result in the lowest average error in the predictions. On the other hand, when the objective is to support "singular, monumental decisions made by businesses, such as how to position a new entrant within a competitive market" \cite{Sanders2017}, the function cannot be completely treated as a black box. In this event, visualizations are and will be a necessary persuasion tool to convey the message of what and why a specific data-driven decision should be made.

%
%
%

\end{document}